\documentclass{article}

\usepackage{PRIMEarxiv}

\usepackage[utf8]{inputenc} 
\usepackage[T1]{fontenc}    
\usepackage{hyperref}       
\usepackage{url}            
\usepackage{booktabs}       
\usepackage{amsfonts}       
\usepackage{nicefrac}       
\usepackage{microtype}      
\usepackage{lipsum}
\usepackage{fancyhdr}       
\usepackage{graphicx}       
\graphicspath{{media/}}     

\usepackage{amsmath}
\usepackage{cleveref}
\usepackage{multirow}
\usepackage{makecell}
\usepackage{amssymb}
\usepackage{pifont}
\usepackage{wrapfig}
\usepackage[dvipsnames,table]{xcolor}

\definecolor{mygray}{gray}{.9}
\definecolor{ggray}{RGB}{127,127,127}
\definecolor{reda}{RGB}{192,0,0}
\definecolor{redb}{RGB}{217,148,143}
\definecolor{myyellow}{RGB}{190,144,0}
\definecolor{mygreen}{RGB}{80,100,40}
\definecolor{myblue}{RGB}{30,90,100}
\definecolor{dark-gray}{gray}{0.20}
\definecolor{middle-gray}{gray}{0.85}
\definecolor{light-gray}{gray}{0.93}
\definecolor{lightblue}{rgb}{0.85, 0.95, 1}
\definecolor{lighterblue}{rgb}{0.9, 0.97, 1}
\definecolor{palestblue}{rgb}{0.95, 0.98, 1}

\title{NavWM: A Unified Navigation World Model for Foresight-Driven Planning}

\author{
  Yanghong Mei\textsuperscript{1, 3}, 
  Longteng Guo\textsuperscript{1}, 
  Ming-Ming Yu\textsuperscript{2}, 
  Guiyu Zhao\textsuperscript{1, 3}, 
  Xingjian He\textsuperscript{1}, 
  Jing Liu\textsuperscript{1, 3}\thanks{Corresponding author: \texttt{jliu@nlpr.ia.ac.cn}} \\
  \vspace{0.1in} \\
  \textsuperscript{1}Institute of Automation, Chinese Academy of Sciences \\
  \textsuperscript{2}Beihang University \\
  \textsuperscript{3}School of Artificial Intelligence, University of Chinese Academy of Sciences \\
}

\begin{document}
\maketitle

\begin{abstract}
Conventional visual navigation policies often struggle with myopic decision-making and mode collapse in complex environments. While world models offer a promising alternative, existing paradigms typically isolate perception, generation, and control, failing to capture their shared spatio-temporal dynamics. In this paper, we propose NavWM, a unified navigation world model that seamlessly integrates latent world reasoning, multimodal action prediction, and controllable visual generation. At its core, NavWM leverages latent world tokens to distill geometric and semantic priors, endowing the agent with robust structural understanding. To overcome the limitations of deterministic policies, we introduce an anchor-based multimodal trajectory forecasting framework that generates a diverse action space. This inherent diversity explicitly empowers the generative world model to act as a robust closed-loop planner, utilizing visual foresight to evaluate and select the optimal path. Extensive experiments across diverse robotics datasets demonstrate that NavWM significantly advances the state-of-the-art, delivering remarkable improvements in both high-fidelity future state generation and zero-shot navigation success.
\end{abstract}

\keywords{World model \and Visual navigation}

\section{Introduction}
\label{sec:intro}
Visual navigation~\cite{shah2022gnm, shah2023vint, sridhar2024nomad, liu2025citywalker} stands as a cornerstone of embodied intelligence, underpinning a wide array of real-world applications such as autonomous driving, automated delivery, and robotic services. It requires an agent to rely solely on camera observations to safely maneuver through arbitrary, unstructured environments, actively avoiding pedestrians and obstacles to reach a designated goal. Achieving this complex task demands far more than static scene comprehension; it fundamentally requires the agent to anticipate and reason about future world states while in motion~\cite{bar2025navigation}.

Conventional visual navigation policies typically learn a direct visuo-motor mapping from visual observations to action sequences, trained on large-scale web videos or embodied trajectories \cite{shah2022gnm, shah2023vint, liu2025citywalker, mei2025urbannav}. However, without explicit foresight, these purely reactive policies often suffer from myopic decision-making and limited generalization to unseen environments \cite{gervet2023navigating, song2025survey}.

\begin{figure}[tb]
  \centering
  \includegraphics[width=16.5cm]{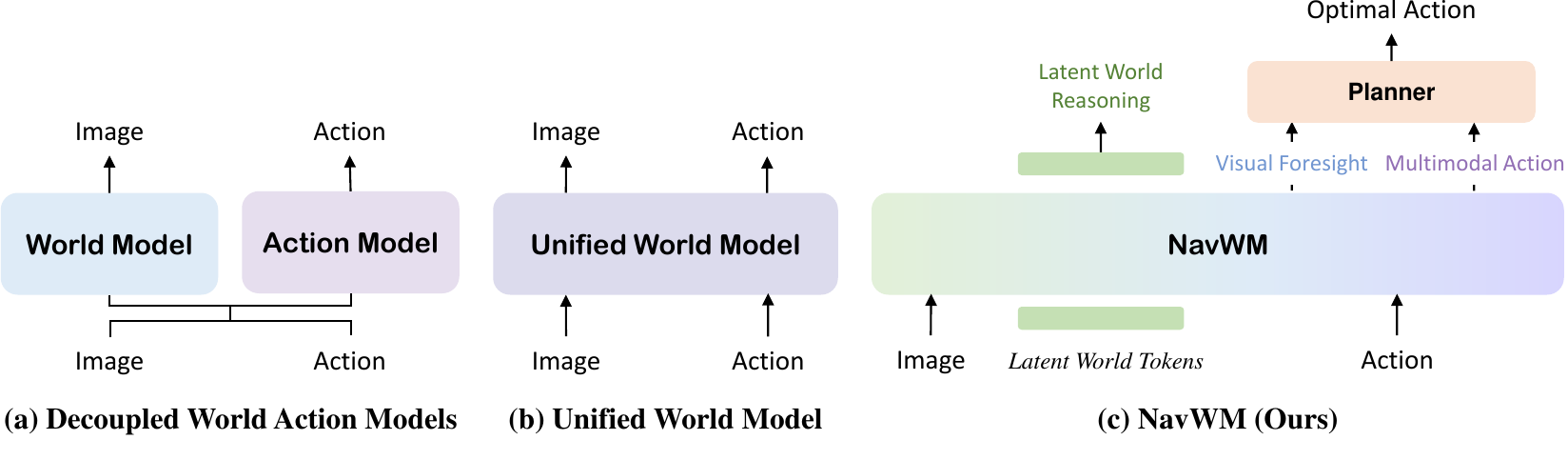}
\caption{\textbf{Comparison of world model paradigms.} 
(a) Decoupled world and action models. 
(b) A unified world model with independent outputs. 
(c) Our NavWM integrates latent world reasoning and multimodal actions, enabling the world model to act as a planner via visual foresight.}
  \label{fig:teaser}
\end{figure}

To address these limitations, recent works have begun incorporating world models for navigation. While promising, existing paradigms still face three key challenges. First, modular pipelines (e.g., \cite{lin2025navcot, bar2025navigation}) typically train the navigation policy and the world model separately (Fig.~\ref{fig:teaser}(a)). This decoupled design fails to leverage their shared spatiotemporal dynamics, resulting in sub-optimal representations. Second, although a few recent works attempt to unify the navigation policy and the world model \cite{hu2025astranav, dong2025unified} (Fig.~\ref{fig:teaser}(b)), their latent representations often lack explicit scene abstractions. Without structured cues about environmental geometry or semantics, the model must infer spatial regularities directly from raw features, which can hinder stable long-horizon prediction and planning. Finally, real-world navigation exhibits inherently multi-modal futures, where multiple feasible trajectories may lead toward the same goal. However, many existing methods model action prediction as a single trajectory, collapsing the multi-modal action space into one mode. Consequently, the predicted actions become overly concentrated, making the agent prone to sub-optimal decisions or stalling in local minima.

To bridge this critical gap, we present a \textbf{Unified Navigation World Model} (\textbf{NavWM}), that seamlessly integrates latent world reasoning, controllable visual generation, and multimodal action prediction \footnote{``Multimodal'' denotes predicting multiple spatial trajectory modes \cite{chi2025diffusion, gu2021densetnt, zhao2021tnt}.}. Unlike prior disjointed paradigms, NavWM is built upon the core insight that perception, generation, and control fundamentally share a reliance on modeling underlying environmental structures and spatio-temporal dynamics. By unifying these heterogeneous tasks within a shared Bidirectional Mamba backbone, our framework explicitly exploits their inherent synergy during the training phase to foster mutually reinforcing representations. This unified learning process, in turn, empowers the inference phase by supporting \textbf{closed-loop planning} via \textbf{visual foresight}. Consequently, the agent can simulate future visual observations for diverse trajectory candidates, enabling it to evaluate goal alignment and filter the optimal path within a simulated ``multiverse'' of possibilities.

To encourage structured scene representations, we introduce dedicated latent world tokens that capture geometric and semantic cues of the environment. These tokens are supervised using depth and semantic signals derived from visual foundation models, guiding the model to learn spatially consistent representations. Such structured cues provide physically grounded context for long-horizon prediction and improve the stability of future observation synthesis.

In addition, real-world navigation often admits multiple feasible future trajectories toward the same goal. However, many prior methods rely on predicting a single trajectory \cite{shah2022gnm, shah2023vint, dong2025unified}, which restricts exploration and can trap the agent in sub-optimal behaviors \cite{chi2025diffusion}. To address this limitation, NavWM adopts an anchor-based multimodal trajectory prediction framework that generates diverse yet physically plausible motion hypotheses. This multimodal proposal mechanism expands the candidate action space, allowing the world model to evaluate alternative futures through visual foresight and select more reliable navigation decisions.

Extensive experimental results demonstrate that NavWM achieves outstanding performance across a diverse range of robotics datasets (Go Stanford \cite{hirose2018gonet}, SCAND \cite{karnan2022socially}, RECON \cite{shah2021rapid}, HuRoN \cite{hirose2023sacson}, and Tartan Drive \cite{triest2022tartandrive}). In offline evaluations for single-step prediction, NavWM significantly improves image generation quality, evidenced by a notable increase in PSNR from 14.17 to 17.34 (\cref{tab:main_results}). In complex, long-horizon navigation tasks, NavWM demonstrates formidable navigation performance alongside robust scene reconstruction capabilities. Notably, it elevates the navigation success rate from 66\% to 72\% and achieves a remarkable 44\% success rate even during zero-shot inference in unseen environments (\cref{tab:main_rollout}, \cref{fig:vis_traj}). Ablation experiments demonstrate that the joint training of reasoning, control, and world modeling yields mutually reinforcing effects, fundamentally driven by their shared reliance on underlying spatio-temporal dynamics. Moreover, our multi-modal action prediction approach elevates trajectory diversity and strengthens the model's exploration capacity, inherently supporting the use of the generative world model as a high-level supervisory planner.

In summary, the main contributions of our work are threefold:
\begin{itemize}
\item We propose NavWM, a unified navigation world model that jointly learns structured perception, controllable world modeling, and multimodal action prediction within a shared architecture.

\item We introduce a world-model-based visual foresight planning mechanism that evaluates candidate trajectories by simulating future observations, enabling goal-aligned closed-loop navigation.

\item Extensive experiments demonstrate that NavWM achieves superior performance in both visual state simulation and complex navigation, significantly boosting generation quality and success rates across diverse environments.
\end{itemize}

\section{Related Work}
\label{sec:rel_work}

\subsection{Visual Navigation}
Visual navigation stands as a fundamental challenge in robotics, necessitating robust capabilities in scene dynamics modeling and strategic planning \cite{chaplot2020learning, ding2025lavira, frey2023fast, yin2025unigoal, CNav2025}. Early works \cite{shah2022gnm, shah2023vint} learn a deterministic end-to-end mappings from current states to actions from real-world navigation data, whereas Sridhar et al. \cite{sridhar2024nomad} adopt diffusion policy to represent stochastic action distributions. Recent approaches \cite{liu2025citywalker, mei2025urbannav} learn generalizable visual navigation models from large-scale web videos. Nevertheless, they suffer from deterministic modeling or overly peaked distributions, leading to mode collapse that severely hinders environmental exploration \cite{gervet2023navigating}. Moreover, they focus predominantly on reactive low-level execution, devoid of hierarchical decision-making capabilities. Conversely, we present a unified framework designed to forecast multimodal trajectories alongside future states. By deploying the world model as a high-level planner, our approach enables the agent to execute more effective and far-sighted exploration.

\subsection{World Models for Navigation}
World models \cite{ha2018world} capture environmental dynamics and temporal evolution by predicting future states. The complex spatio-temporal dynamics of navigation make it an ideal downstream testbed for world models \cite{frey2023fast, dong2025unified}. Early diffusion-based works \cite{assran2023self, brooks2024video, bruce2024genie, agarwal2025cosmos} have demonstrated remarkable capabilities in high-fidelity simulation. Other works actively exploring viewpoint-conditioned future state generation \cite{feng2024i2vcontrol, he2024cameractrl, he2025cameractrl}. Recent works have explored leveraging world models as high-level planners, integrating them into navigation rollouts to explicitly guide action policies \cite{yao2025navmorph, bar2025navigation, hu2025astranav}. Nevertheless, existing approaches often fail to exploit the underlying capabilities inherently shared by both navigation policies and generative world models: namely, deep scene comprehension and spatio-temporal dynamics modeling. To bridge this gap, this paper proposes a unified framework that explicitly incorporates a latent world reasoning task. By employing a joint training strategy, we enable scene perception, the navigation policy, and the world model to mutually reinforce one another during the learning process.

\section{Methodology}
\label{sec:method}

\subsection{Problem Formulation}
\label{subsec:formulation}
Given the current world state, historical observation context, and navigation targets (e.g., Image Goal), NavWM is designed to learn the underlying environmental dynamics. It concurrently predicts the optimal action and the subsequent state transition. Notably, this predictive capability extends to target-agnostic scenarios, allowing the model to imagine future spatiotemporal evolutions under different hypothetical actions.
Formally, given the current egocentric RGB observation $o_t \in \mathbb{R}^{H \times W \times 3}$, a historical context $\mathcal{O}_c = \{o_{t-M}, \dots, o_{t-1}\}$ within a temporal window $M$, and a goal image $o_g$, our objective is to learn a unified model $F_\theta$ that approximates the joint conditional distribution of multimodal trajectories $\mathcal{T} = \{\hat{a}_{t:t+N-1}^{(k)}\}_{k=1}^K$ and the corresponding future observations  $\mathcal{O} = \{\hat{o}_{t+1:t+N}^{(k)} \}_{k=1}^K$ over a prediction horizon $N$:
\begin{equation}
    (\mathcal{T}, \mathcal{O}) \sim F_{\theta}(\cdot | \mathcal{O}_c, o_t, o_g),
\end{equation}
where the action $a = (\mu, \phi)$ consists of a translational component $\mu \in \mathbb{R}^2$ and a rotational component $\phi \in \mathbb{R}$, defined in the agent's egocentric coordinate frame.

\begin{figure}[tb]
  \centering
  \includegraphics[width=16.5cm]{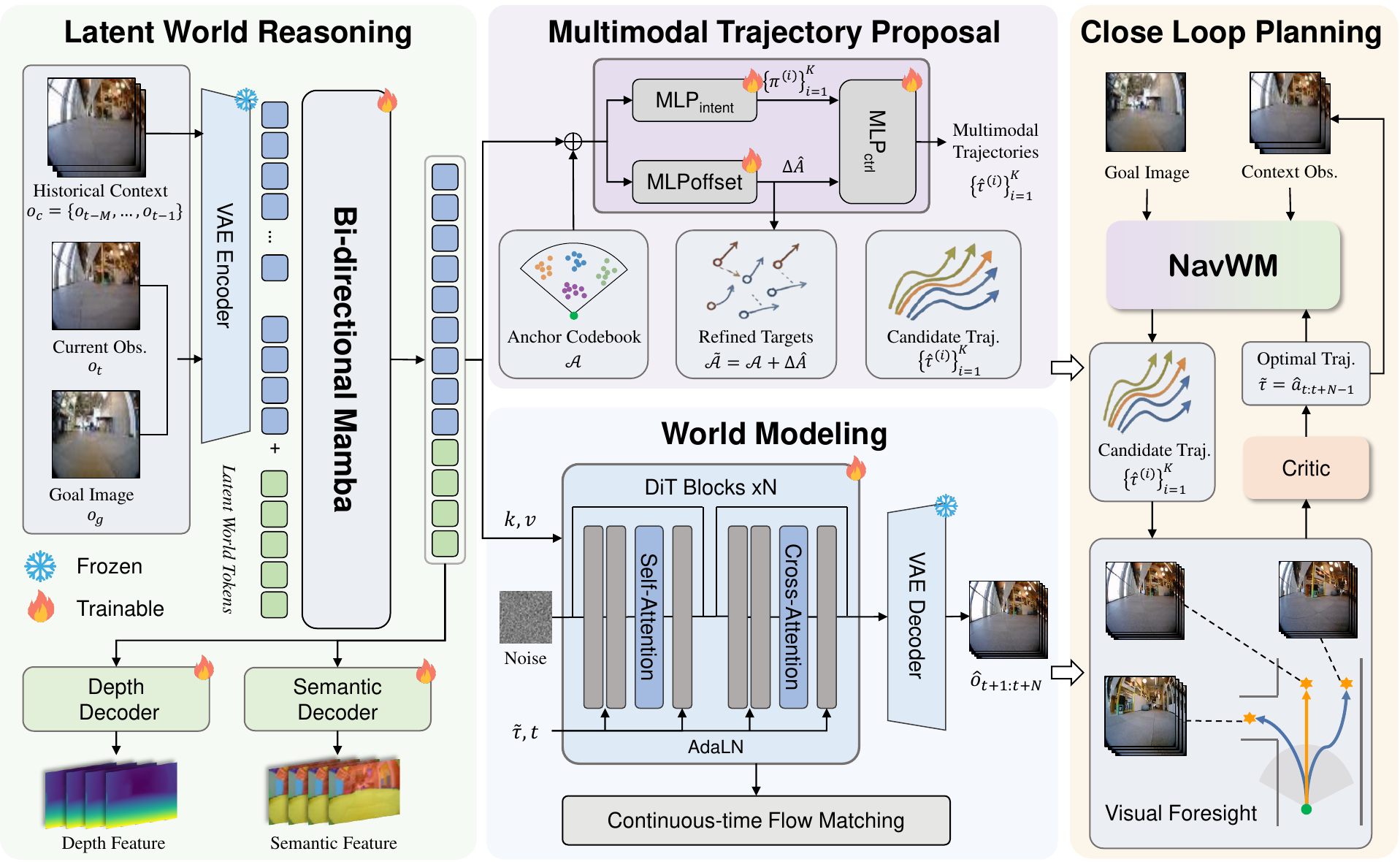}
  \caption{
    \textbf{Overview of the NavWM framework.} The unified navigation world model learns latent scene abstractions from historical observations and goals via latent world reasoning, supervised by depth and semantic predictions. It then proposes multimodal trajectory hypotheses and predicts future observations through world modeling, enabling closed-loop navigation planing via visual foresight. 
    }
  \label{fig:pipeline}
\end{figure}

\subsection{NavWM Overview}
\label{subsec:overview}
As illustrated in Figure \ref{fig:pipeline}, NavWM employs a State Space Model (SSM) as the backbone, which branches into three specialized heads dedicated to latent world reasoning, action prediction and future synthesis, respectively. The historical frames $\mathcal{O}_c$, current observation $o_t$, and goal $o_g$ are first projected into a compact latent space via a pretrained VAE encoder \cite{blattmann2023stable}. Subsequently, these latent embeddings are prepended with a sequence of learnable \textit{Latent World Tokens} and fed into a Bidirectional Mamba backbone \cite{gu2024mamba} for spatiotemporal encoding. We employ an Attentional Pooling Layer to compress the encoded historical tokens into a fixed-length representation. This simple yet effective strategy significantly enhances computational efficiency for long-horizon inference.

\subsection{Latent World Reasoning}
\label{subsec:world_reasoning}
To internalize the 3D geometry and semantics essential for foresighted navigation, NavWM introduces a \textbf{Latent World Reasoning} mechanism that constructs explicit scene abstractions to ground subsequent planning and future imagination. Specifically, we introduce learnable \textit{Latent World Tokens} to assimilate these environmental priors. For robust supervision, we distill knowledge from visual foundation models (Depth Anything V2 \cite{yang2024depth} for geometry and SAM \cite{kirillov2023segment} for semantics) to generate dense future pseudo-labels, $\mathcal{D}_{gt} = \{d_n\}$ and $\mathcal{S}_{gt} = \{s_n\}$, over the horizon $n \in \{t+1, \dots, t+N\}$. 
Following spatiotemporal encoding, these Latent World Tokens are spatially upsampled via a CNN decoder to forecast the scene abstractions, yielding predictions $\hat{\mathcal{D}} = \{\hat{d}_n\}$ and $\hat{\mathcal{S}} = \{\hat{s}_n\}$.

To mitigate the severe scale ambiguity inherent in diverse indoor and outdoor navigation environments, we supervise the geometric predictions using a Scale-Invariant Loss \cite{eigen2014depth}, defined as:
\begin{equation}
\mathcal{L}_{si} = \frac{1}{P} \sum_{p=1}^P g_p^2 - \frac{\lambda}{P^2} \left( \sum_{p=1}^P g_p \right)^2,
\label{eq:si_loss}
\end{equation}
where $g_p = \log \hat{d}_p - \log d_p$, $p$ indexes the valid pixels within a single frame, and $P$ denotes the total number of pixels. Concurrently, the semantic feature alignment is supervised via a standard Mean Squared Error (MSE) loss. 
The final Latent World Reasoning objective is formulated as the weighted sum of these two components, averaged over the temporal prediction horizon:
\begin{equation}
\mathcal{L}_{reason} = \lambda_{depth} \mathcal{L}_{si}(\mathcal{D}_{gt}, \hat{\mathcal{D}}) + \lambda_{sem} \mathcal{L}_{mse}(\mathcal{S}_{gt}, \hat{\mathcal{S}}).
\end{equation}
By explicitly optimizing this auxiliary objective, the model effectively internalizes the underlying geometric and semantic priors of the scene.

\subsection{Multimodal Trajectory Proposal}
\label{subsec:multimodal_traj}

Reliable foresight planning requires exploring multiple plausible future actions rather than committing to a single deterministic trajectory, which is prone to sub-optimal local minima. To provide the world model planner with a diverse and expressive action space, NavWM adopts a multimodal trajectory proposal module that predicts a set of physically plausible future motion hypotheses. 

Following prior work on trajectory prediction \cite{gu2021densetnt}, we model future uncertainty as arising from two complementary factors: \emph{intent uncertainty}, corresponding to the high-level navigation target, and \emph{control uncertainty}, reflecting low-level motion variations toward that target. Accordingly, the trajectory proposal process is decomposed into two stages: anchor-based target prediction and target-conditioned trajectory generation.

\subsubsection{Anchor-based Target Prediction.}
We first predict a set of candidate navigation targets represented as relative 2D positions in the agent's egocentric coordinate frame. Specifically, we construct an anchor codebook $\mathcal{A} \in \mathbb{R}^{K \times 2}$ that captures the most representative target locations over horizon $N$. The anchors are obtained offline by applying K-Means clustering followed by Farthest Point Sampling on ground-truth target coordinates. Given the scene representation $\mathcal{H}$ produced by the world model encoder, we estimate the likelihood of each anchor and its corresponding spatial refinement. Concretely, the anchor coordinates are concatenated with the scene features and passed through two lightweight MLP heads: an intent head that predicts the anchor probabilities $\hat{\boldsymbol{\pi}} \in \mathbb{R}^K$, and a control head that regresses continuous spatial offsets $\Delta \hat{\mathbf{A}} \in \mathbb{R}^{K \times 2}$:
\begin{equation}
    \hat{\boldsymbol{\pi}} = \text{Softmax}(\text{MLP}_{intent}([\mathcal{H} \parallel \mathcal{A}])), 
    \quad
    \Delta \hat{\mathbf{A}} = \text{MLP}_{control}([\mathcal{H} \parallel \mathcal{A}]).
\end{equation}
The refined targets are therefore expressed as $\widetilde{\mathcal{A}} = \mathcal{A} + \Delta \hat{\mathbf{A}}$. 
During training, we identify the anchor $k^*$ that best matches the ground-truth and optimize a joint objective comprising an intent classification loss and an offset regression loss:
\begin{equation}
    \mathcal{L}_{anchor} =
    -\log(\hat{\boldsymbol{\pi}}_{k^*})
    +
    \mathcal{L}_{huber}(\Delta \hat{x}_{k^*}, \Delta x_{gt})
    +
    \mathcal{L}_{huber}(\Delta \hat{y}_{k^*}, \Delta y_{gt}),
\end{equation}
where $(\Delta x_{gt}, \Delta y_{gt})$ is the ground-truth offset relative to the matched anchor.

\subsubsection{Target-Conditioned Trajectory Prediction.}
Given the refined targets $\widetilde{\mathcal{A}}$, we further predict a full sequence of future waypoints for each candidate target. Instead of autoregressively generating the trajectory step-by-step, which incurs significant computational overhead, we adopt a direct prediction strategy that assumes conditional independence across future timesteps \cite{cui2019multimodal, chai2019multipath, mangalam2020not, zhao2021tnt}. 
Specifically, the scene features $\mathcal{H}$ and refined targets $\widetilde{\mathcal{A}}$ are fed into an MLP to directly regress the waypoint sequence $\hat{\mathcal{T}}_k$ associated with each candidate target. To avoid the non-differentiable $\textit{argmax}$ operation during training, supervision is applied only to the trajectory corresponding to the matched anchor $k^*$. The trajectory loss is defined as:
\begin{equation}
    \mathcal{L}_{traj} = \mathcal{L}_{huber}(\hat{\mathcal{T}}_{k^*}, \mathcal{T}_{gt}).
\end{equation}
The final action proposal objective combines the anchor prediction and trajectory regression losses:
\begin{equation}
    \mathcal{L}_{action} = \mathcal{L}_{anchor} + \lambda_{traj}\mathcal{L}_{traj},
\end{equation}
where $\lambda_{traj}$ balances the two components. The resulting module produces a diverse set of trajectory candidates that serve as action hypotheses for subsequent visual foresight evaluation.

\subsection{World Modeling for Visual Foresight}
\label{world_model}

To enable foresight-based planning, NavWM learns a world model that predicts future visual observations conditioned on candidate trajectories. Given the scene representation and predicted motion, the model simulates how the environment would evolve under different action hypotheses.
As illustrated in Figure~\ref{fig:pipeline}, the input visual latents and predicted actions are processed by a stack of Conditional Diffusion Transformer (CDiT) blocks \cite{bar2025navigation}, which provide a more efficient alternative to directly applying a vanilla Diffusion Transformer (DiT) \cite{peebles2023scalable}. To explicitly encode camera motion, the predicted actions are first transformed into dense optical flow fields \cite{sitzmann2021light}, denoted as $\mathcal{F}$. The motion representation is spatially pooled and fused with the diffusion timestep embedding, forming a joint conditioning signal that modulates intermediate activations through adaptive layer normalization (AdaLN) \cite{xu2019understanding}. 

Meanwhile, the scene features $\mathcal{H}$ produced by the unified world model backbone are injected via cross-attention, providing contextual guidance for future synthesis. The resulting latent tokens are then unpatchified into spatial latent maps and decoded by the VAE decoder to generate realistic future observations.

We optimize the NavWM world model via continuous-time Flow Matching \cite{liu2022flow, lipman2022flow}, formulating future synthesis as a deterministic mapping from a Gaussian prior to the target latent distribution. 
Specifically, given a ground-truth latent $z_1$ and a noise sample $z_0 \sim \mathcal{N}(0, \mathbf{I})$, we define a probability path via linear interpolation $z_{\tau} = \tau z_1 + (1 - \tau) z_0$, where $\tau \sim \mathcal{U}(0, 1)$. The network $v_\theta$ is trained to regress the constant vector field $u_{\tau} = z_1 - z_0$ that drives the flow from noise to data, supervised by the following objective:
\begin{equation}
    \mathcal{L}_{visual} = \mathbb{E}_{\tau \sim \mathcal{U}(0,1), z_0 \sim \mathcal{N}, z_1 \sim q} \left[ \left\| v_\theta(z_\tau, \tau, \mathcal{F}, \mathcal{H}) - (z_1 - z_0) \right\|_2^2 \right],
\label{eq:fm_loss}
\end{equation}
This formulation ensures straighter synthesis trajectories and highly efficient sampling during inference.

\begin{table}[t]
\renewcommand{\cellalign}{l}
\caption{\textbf{Offline evaluation.} Average performance of action prediction and visual generation across four primary datasets.}
\label{tab:main_results}
\centering
\scalebox{1.0}{
\setlength{\tabcolsep}{5pt}
\begin{tabular}{l cccc cccc}
\toprule
\multirowcell{2}{\textbf{Method}} & \multicolumn{4}{c}{\textbf{Navigation}} & \multicolumn{4}{c}{\textbf{Visualization}} \\
\cmidrule(lr){2-5} \cmidrule(lr){6-9}
& {\bf ATE} $\downarrow$ & {\bf RPE} $\downarrow$ & {\bf AOE} $\downarrow$ & {\bf MAOE} $\downarrow$ & {\bf SSIM} $\uparrow$ & {\bf PSNR} $\uparrow$ & {\bf LPIPS} $\downarrow$ &  {\bf DreamSim} $\downarrow$ \\
 
\midrule
GNM \cite{shah2022gnm} & 1.519 & 0.622 & 18.165 & 23.961 & - & - & - & - \\
ViNT \cite{shah2023vint} & 1.424 & 0.617 & 17.074 & 23.831 & - & - & - & - \\

NoMaD \cite{sridhar2024nomad} & 1.070 & 0.376 & 12.463 & 17.645 & - & - & - & - \\

Diamond \cite{alonso2024diffusion} & - & - & - & - & 0.296 & 8.857 & 0.425 & 0.161 \\
Anloe-7B \cite{chern2024anole} & 1.735 & 0.749 & 18.288 & 24.878 & 0.288 & 9.176 & 0.513 & 0.163 \\

NWM \cite{bar2025navigation} & 0.642 & 0.211 & 10.496 & 16.534 & 0.422 & 14.343 & 0.295 & 0.091 \\

UniWM \cite{dong2025unified} & 0.302 & 0.116 & 9.468 & 13.221 & 0.435 & 14.172 & 0.282 & 0.102 \\
\rowcolor{lightblue} {\bf NavWM (Ours)} & {\bf 0.207} & {\bf 0.066} & {\bf 8.152} & {\bf 12.855} & {\bf 0.507} & {\bf 17.340} & {\bf 0.243} & {\bf 0.084} \\
\bottomrule
\end{tabular}
}
\end{table}

\subsection{Two-Stage Joint Training}
\label{subsec:joint_training}
NavWM is trained with a two-stage optimization strategy to leverage the complementary strengths of perception, control, and visual foresight while mitigating the train–inference discrepancy. In the first stage, all components are jointly optimized using teacher forcing. Ground-truth trajectories are provided to the CDiT world modeling module, allowing the perception, action prediction, and visual synthesis branches to learn mutually consistent spatiotemporal representations. The second stage addresses the exposure bias caused by the mismatch between training and inference. Specifically, we freeze the SSM backbone and trajectory proposal heads, and fine-tune the CDiT conditioned on trajectories sampled from the predicted distribution rather than the ground truth. This stage improves the robustness of the world model under imperfect action predictions.

To balance the heterogeneous gradients arising from different learning objectives, we adopt the uncertainty-weighted multi-task formulation proposed in~\cite{kendall2018multi}. The overall training objective is defined as
\begin{equation}
\mathcal{L} =
\sum_{k \in \mathcal{T}}
\left(
\frac{1}{2}\exp(-s_k)\mathcal{L}_k + \frac{1}{2}s_k
\right),
\end{equation}
where $\mathcal{T} = \{\text{visual}, \text{action}, \text{reason}\}$ denotes the set of training tasks, and $s_k$ represents the learnable log-variance associated with task $k$. This adaptive weighting mechanism automatically balances the relative contributions of different objectives and eliminates the need for manual loss weighting.

\section{Experiments and Results}
\label{sec:experiment}

\subsection{Experimental Setting}
\label{subsec:setting}

\subsubsection{Datasets.} We train NavWM using a mixture of robotics datasets that cover both indoor and outdoor environments (Go Stanford \cite{hirose2018gonet}, SCAND \cite{karnan2022socially}, RECON \cite{shah2021rapid}, and HuRoN \cite{hirose2023sacson}).Containing precise real-world robot location and rotation, they can be easily converted into navigation actions (i.e., relative camera pose transformations). Paired with corresponding RGB observations, they provide a suitable resource for training navigation policies and world models. Following prior works \cite{sridhar2024nomad, bar2025navigation, dong2025unified}, we discard trajectories shorter than three steps and backward movements, partitioning the remaining data into semantically coherent segments based on the continuous video stream. Additionally, we incorporate the TartanDrive \cite{triest2022tartandrive} as a zero-shot testbed for unseen environments.

\begin{figure}[tb]
  \centering
  \includegraphics[width=16.5cm]{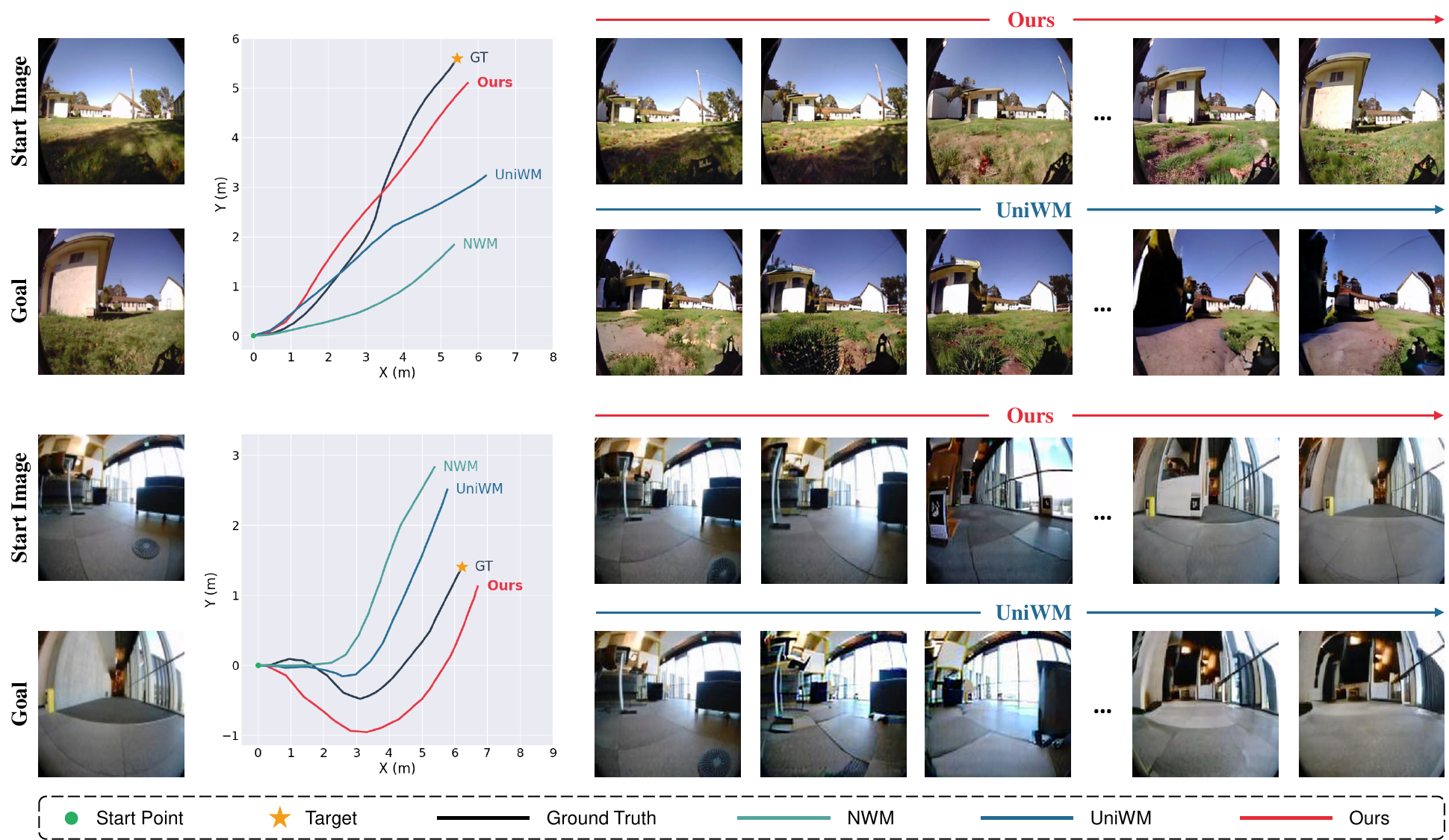}
  \caption{\textbf{Qualitative comparison across indoor and outdoor environments.} Left: Trajectory plots illustrating the alignment of predicted paths from different methods with the Ground Truth (GT). Right: Visual generation results of camera viewpoint control during navigation.}
  \label{fig:vis_traj}
\end{figure}

\subsubsection{Evaluation Metrics.}
For the offline evaluation of the navigation policy, we employ Absolute Trajectory Error (ATE) and Relative Pose Error (RPE) to assess pose consistency \cite{sturm2012evaluating}, while utilizing Average Orientation Error (AOE) and Maximum Average Orientation Error (MAOE) to quantify overall trajectory similarity \cite{liu2025citywalker}. Regarding the generative performance of the world model, pixel-level synthesis quality and structural similarity are measured via PSNR and SSIM \cite{wang2004image}, whereas LPIPS \cite{zhang2018unreasonable} and DreamSim \cite{fu2023dreamsim} are adopted to evaluate perceptual similarity in the latent space.

\subsubsection{Implementation Details.}
In the default setup, the model operates at a resolution of $256 \times 256$, receiving a context of 4 past frames to forecast the waypoints and visual states of the next 4 frames. We optimize 1.5B-parameter NavWM of using the AdamW optimizer \cite{loshchilov2017decoupled}, employing a learning rate schedule that warms up from $5 \times 10^{-5}$ to a peak of $1 \times 10^{-4}$ over the first $500$ steps, followed by a cosine decay to a minimum of $1 \times 10^{-5}$. Using 8 NVIDIA A100 GPUs with a global batch size of 64, we employ teacher forcing in the first stage to train the future synthesis head with ground-truth trajectories for 100,000 steps, applying a goal mask with a $15\%$ probability. In the second stage, the future synthesis head is exclusively fine-tuned for 50,000 steps, conditioned instead on the predicted trajectories.

\subsection{Offline Evaluation}
\label{subsec:offline_eval}

\subsubsection{Baselines.}
For trajectory prediction, we select conventional navigation policies ViNT \cite{shah2023vint} and NoMaD \cite{sridhar2024nomad} as our baselines. ViNT employs a Transformer architecture to learn an end-to-end visuo-motor mapping, whereas NoMaD utilizes a diffusion head to output a non-deterministic policy. Regarding visualization performance, we benchmark against Diamond \cite{alonso2024diffusion} alongside state-of-the-art methods NWM \cite{bar2025navigation} and UniWM \cite{dong2025unified}. NWM is a DiT-based world model; to evaluate its navigation metrics, we leverage its Standalone Planning capability to guide NoMaD. UniWM, fine-tuned from the unified multimodal model Anole \cite{chern2024anole}, serves as an autoregressive model tailored for navigation and visualization.

\subsubsection{Comparison with SOTA Methods.}
\cref{tab:main_results} details the offline evaluation. To assess the fundamental navigation performance independently of the generative planner, we perform random sampling across the multimodal trajectory predictions to evaluate their baseline accuracy. As shown, NavWM consistently surpasses both the NWM-guided NoMaD and the current state-of-the-art method, UniWM, establishing superior trajectory prediction accuracy (ATE: 0.302 $\to$ 0.207, RPE: 0.116 $\to$ 0.066). Moreover, our visual generation quality outperforms both the diffusion-based NWM and the autoregressive UniWM in terms of structural and perceptual similarity (PSNR: 14.343 $\to$ 17.340, LPIPS: 0.282 $\to$ 0.243), emerging as the new state-of-the-art.

\subsection{Navigation Rollout}
\label{subsec:rollout}
In this phase, we evaluate the model's performance on the Image Goal Navigation task. This requires the model to navigate to the goal image's location relying solely on the initial observation, achieved through an interleaved process of action prediction and visual forecasting \cite{dong2025unified}. We segment the dataset trajectories based on semantic coherence, yielding an average segment length of 43 steps. Navigation is deemed successful when the agent arrives within a 0.5m radius of the goal, based on which we calculate the Success Rate (SR). Additionally, we measure trajectory consistency using ATE and RPE.

\begin{table}[tb]
\renewcommand{\cellalign}{l}
\caption{\textbf{Navigation rollout results.} This table reports various metrics for Image Goal Navigation in both Seen and Unseen environments.} 
\label{tab:main_rollout}
\centering
\scalebox{1.0}{
\setlength{\tabcolsep}{7pt}
\begin{tabular}{l ccc ccc}
\toprule
\multirowcell{2}{{\bf Method}} & \multicolumn{3}{c}{{\bf Seen}} & \multicolumn{3}{c}{{\bf Unseen}} \\
\cmidrule(lr){2-4} \cmidrule(lr){5-7}
& {\bf SR} $\uparrow$ & {\bf ATE} $\downarrow$ & {\bf RPE} $\downarrow$ & {\bf SR} $\uparrow$ & {\bf ATE} $\downarrow$ & {\bf RPE} $\downarrow$ \\
\midrule
ViNT \cite{shah2023vint} & 0.24 & 1.948 & 0.817 & 0.11 & 2.423 & 0.812 \\
NoMaD \cite{sridhar2024nomad} & 0.29 & 1.711 & 0.601 & 0.16 & 2.242 & 0.789 \\
ANOLE-7B \cite{chern2024anole} & 0.16 & 2.237 & 0.911 & 0.11 & 2.334 & 0.932 \\
NWM \cite{bar2025navigation} & 0.58 & 1.141 & 0.360 & 0.23 & 1.712 & 0.647 \\
UniWM \cite{dong2025unified} & 0.66 & 0.645 & 0.229 & 0.36 & 1.122 & 0.375 \\
\rowcolor{lightblue} {\bf NavWM (Ours)} & {\bf 0.72} & {\bf 0.381} & {\bf 0.161} & {\bf 0.44} & {\bf 0.765} & {\bf 0.260} \\
\bottomrule
\end{tabular}
}
\end{table}

\subsubsection{Navigation Performance.}
As shown in \cref{tab:main_rollout}, NavWM achieves superior performance in navigation tasks across both seen and unseen environments. Specifically, when performing random sampling on multimodal trajectory outputs, NavWM attains a navigation success rate of 0.63 in Seen environments, establishing a strong baseline. Crucially, when leveraging the world model as a planner to supervise action generation, NavWM achieves state-of-the-art results in both seen (SR: 0.74) and unseen (SR: 0.44) settings, while also delivering optimal trajectory quality.

\subsubsection{Quantitative Analysis.}
\cref{fig:vis_traj} visualizes the navigation trajectories and corresponding visual generation results for different methods. As observed, NavWM achieves superior alignment with ground truth trajectories. By leveraging diverse trajectory outputs in synergy with a world model planner, NavWM can correct errors in real-time, whereas other methods suffer severely from cumulative drift. Furthermore, visual inspection reveals that while competing methods tend to lose track of the target over long-horizon trajectories, NavWM maintains robustness by consistently reconstructing the target state.

\begin{table}[t]
\caption{\textbf{Module-level ablation study.} Effects of latent World Reasoning (WR), Action Prediction (AP), and World Model (WM).}
\label{tab:module_ablations}
\centering
\scalebox{1.0}{
\setlength{\tabcolsep}{6pt}
\begin{tabular}{ccc cc cccc}
\toprule
\multirowcell{2}{{\bf WR}} & \multirowcell{2}{{\bf AP}} & \multirowcell{2}{{\bf WM}} & \multicolumn{2}{c}{{\bf Navigation}} & \multicolumn{4}{c}{{\bf Visualization}} \\
\cmidrule(lr){4-5} \cmidrule(lr){6-9}
& & & {\bf ATE} $\downarrow$ & {\bf RPE} $\downarrow$ & {\bf SSIM} $\uparrow$ & {\bf PSNR} $\uparrow$ & {\bf LPIPS} $\downarrow$ & {\bf DreamSim} $\downarrow$ \\
\midrule
 & \checkmark & & 0.513 & 0.187 & - & - & - & \\
 & \checkmark & \checkmark & 0.338 & 0.137 & 0.414 & 14.286 & 0.249 & 0.087 \\
\checkmark & & \checkmark & - & - & 0.482 & 16.627 & 0.293 & 0.112 \\
\rowcolor{lightblue} \checkmark & \checkmark & \checkmark & {\bf 0.254} & {\bf 0.113} & {\bf 0.516} & {\bf 17.622} & {\bf 0.240} & {\bf 0.086} \\
\bottomrule
\end{tabular}
}
\end{table}

\begin{table}[h]
\renewcommand{\cellalign}{l}
\caption{\textbf{Effectiveness of multimodal action and world model guidance.} We compare our anchor-based trajectory generation approach against Gaussian Mixture Modeling (GMM) and the diffusion-based head under raw sampling ($\times$) and world model planning (\checkmark).}
\label{tab:multimodal_ablations}
\centering
\scalebox{1.0}{
\setlength{\tabcolsep}{6pt}
\begin{tabular}{lc ccccc}
\toprule
{\bf Methods} & {\bf WM Planning} & {\bf APD $\uparrow$} & {\bf ATE $\downarrow$} & {\bf RPE $\downarrow$}  & {\bf AOE $\downarrow$} & {\bf MAOE $\downarrow$} \\
\midrule
\multirowcell{2}{GMM} & $\times$ & - & 0.88 & 0.26 & 12.03 & 17.54 \\
& \checkmark & 0.31 & 0.79 & 0.23 & 11.62 & 16.84 \\
\midrule
\multirowcell{2}{Diffusion} & $\times$ & - & 0.86 & 0.25 & 11.79 & 17.18 \\
& \checkmark & 0.59 & 0.58 & 0.18 & 10.56 & 14.50 \\
\midrule
\multirowcell{2}{Anchor-based (Ours)} & $\times$ & - & 0.85 & 0.24 & 11.33 & 16.86 \\
& \checkmark & {\bf 1.49} & {\bf 0.36} & {\bf 0.14} & {\bf 8.03} & {\bf 12.34} \\

\bottomrule
\end{tabular}
}
\end{table}

\subsection{Ablations}
\label{subsec:ablations}
We conduct ablation studies on the combined RECON and HuRoN datasets to investigate the following questions:
\begin{itemize}
    \item \textbf{Q1.} Do perception, action, and vision truly synergize in a unified framework?
    \item \textbf{Q2.} How effective is the anchor-based multimodal trajectory prediction?
    \item \textbf{Q3.} Does employing a world model as a generative planner effectively enhance trajectory quality?
    \item \textbf{Q4.} To what extent does increasing candidate count improve performance?
\end{itemize}

\begin{wrapfigure}{r}{0.5\textwidth}
  \vspace{-5mm}
  \centering
  \includegraphics[width=0.95\linewidth]{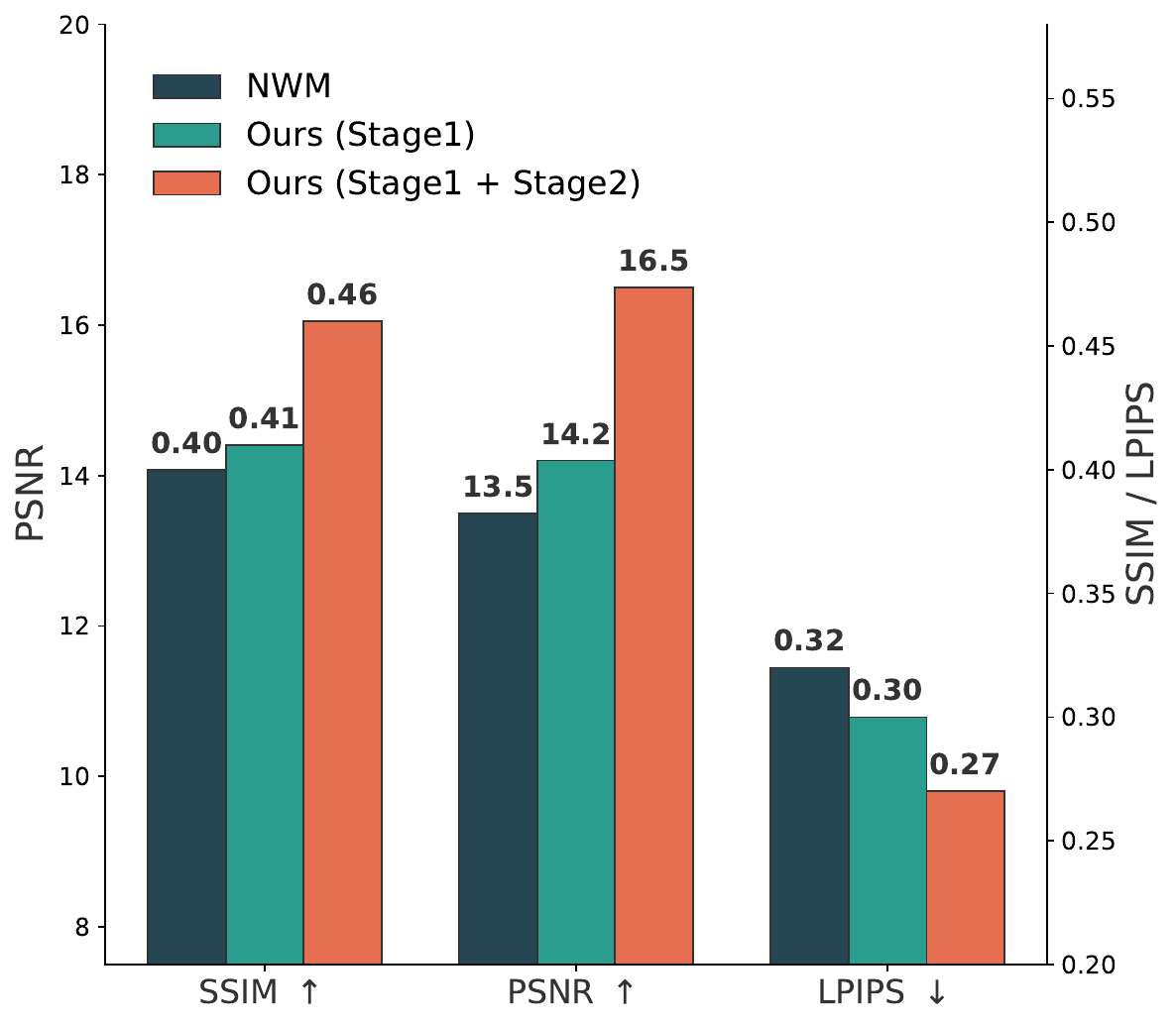}
  \caption{\textbf{Training stage ablation.} Performing stage two fine-tuning significantly improves image reconstruction quality.}
  \label{fig:stage_ablation}
  \vspace{-5mm}
\end{wrapfigure}

\subsubsection{Unified Framework vs. Decoupled Modules.}
To answer \textbf{Q1}, we conduct module ablation study on the latent world reasoning, action prediction, and world model, with the quantitative results summarized in \cref{tab:module_ablations}. To assess the raw multimodal policy, we compute metrics directly on probabilistically sampled trajectories without world model planning. As shown, integrating world reasoning and visual foresight drastically enhances trajectory quality (ATE 0.513 $\to$ 0.164 and RPE 0.187 $\to$ 0.113). Conversely, the inclusion of scene abstractions and action prediction respectively boosts the structural (PSNR 14.286 $\to$ 17.622) and perceptual (LPIPS 0.293 $\to$ 0.240) fidelity of the generated images. Furthermore, our training stage ablation demonstrates that two-stage fine-tuning yields significant improvements in overall image reconstruction quality (\cref{fig:stage_ablation}). These results validate the efficacy of our unified framework and training paradigm.

\subsubsection{Multimodal Diversity and Feasibility.}
Addressing \textbf{Q2}, we compare our anchor-based multimodal trajectory prediction method against Gaussian Mixture Modeling (GMM) and the diffusion-based NoMaD. Additionally, we introduce Average Pairwise Distance (APD) to explicitly evaluate trajectory diversity. As shown in \cref{tab:multimodal_ablations}, under raw probabilistic sampling without the planner, NavWM achieves the best navigation performance while concurrently maintaining the highest APD. This proves that our method effectively mitigates the mode collapse inherent in prior baselines, simultaneously maximizing topological diversity and physical feasibility to significantly boost exploration capabilities without compromising accuracy.

\subsubsection{Efficacy of World Model Planning.}
We compare trajectories generated via raw probabilistic sampling against those refined by the world model planner, thereby answering \textbf{Q3}. As demonstrated in \cref{tab:multimodal_ablations}, supervision from the NavWM world model effectively enhances navigation performance. Furthermore, we observe that these performance gains are significantly more pronounced for our anchor-based multimodal prediction compared to the GMM and diffusion-based NoMaD baselines. This proves that the trajectories generated by NavWM achieve superior environmental exploration, naturally providing a highly compatible action space for high-level planners.

\begin{figure}[tb]
  \centering
  \includegraphics[width=16.5cm]{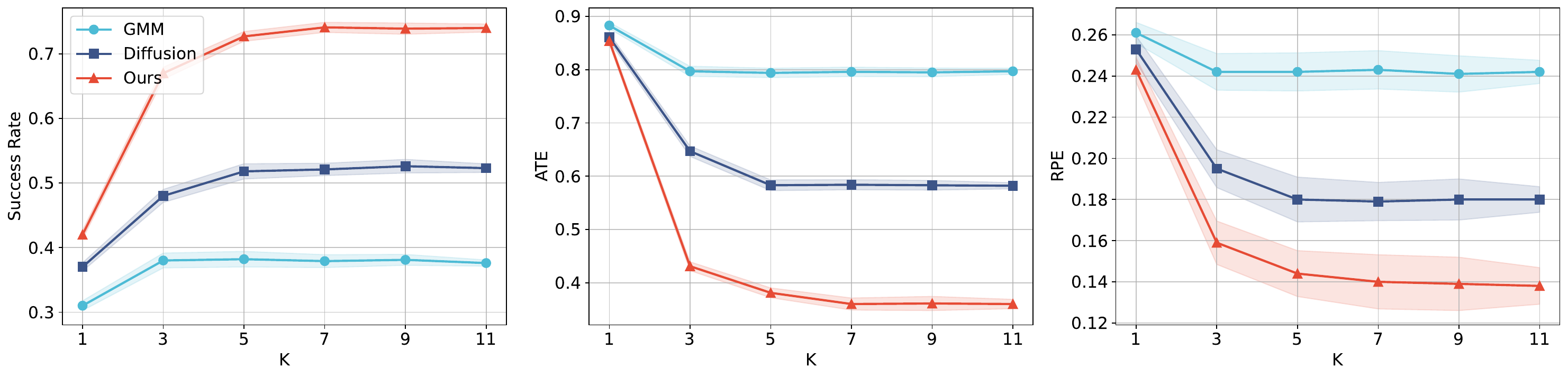}
  \caption{\textbf{Navigation performance vs. candidate trajectory count ($K$).} NavWM shows sustained gains up to $K \approx 7$, outperforming Diffusion ($K \approx 5$) and GMM (earliest saturation), balancing diversity and accuracy.}
  \label{fig:K_ablation}
\end{figure}

\subsubsection{Impact of Candidate Trajectory Count.}
To address \textbf{Q4}, we ablated the number of candidate trajectories $K$ to analyze its effect on navigation metrics under the condition that the world model acts as the planner. As shown in \cref{fig:K_ablation}, navigation performance improves as $K$ increases, which is most evident in NavWM. However, this improvement eventually saturates at a higher $K$. Specifically, the performance plateaus at approximately 7 for NavWM and 5 for the diffusion-based strategy, while the GMM reaches saturation at an even lower value. This implies that NavWM's multimodal predictions maintain high fidelity while offering greater diversity, facilitating more efficient environmental exploration and allowing the system to benefit from a larger set of candidates.

\section{Conclusion}
\label{conclusion}
We present NavWM, a unified world model that overcomes the limitations of deterministic navigation policies by jointly optimizing latent scene reasoning, multimodal action prediction, and visual generation. By distilling geometric and semantic priors through Latent World Tokens, NavWM achieves robust structural understanding. Crucially, our multimodal trajectory framework generates a diverse action space that circumvents local minima and empowers the generative world model to perform optimal closed-loop planning via visual foresight. Extensive evaluations confirm that NavWM significantly advances the state-of-the-art in both high-fidelity future simulation and zero-shot navigation, establishing a powerful new paradigm for unifying perception, generation, and control.

\section{Acknowledgments}
This research is supported by Artificial Intelligence-National Science and Technology Major Project (2023ZD0121200), the National Natural Science Foundation of China (62531026, 62437001, 62436001), the Natural Science Foundation of Jiangsu Province under Grant BK20243051 and the Strategic Priority Research Program of Chinese Academy of Sciences under Grant XDB1350103.

\bibliographystyle{unsrt}  
\bibliography{references}  

\clearpage

\end{document}